\documentclass{article}

\usepackage[preprint]{neurips_2021}

\usepackage{microtype}
\usepackage{graphicx}
\usepackage{subfig}
\usepackage{booktabs} %

\usepackage{siunitx} %

\usepackage{amsmath} %
\usepackage{hyperref} %
\usepackage{listings} %
\usepackage{xcolor} %
\usepackage{hhline} %

\usepackage[para]{footmisc} %

\usepackage{amsthm} %
\usepackage{amsfonts} %
\usepackage{dsfont} %

\usepackage{wrapfig}

\usepackage{hyperref}

\title{A Large Batch Optimizer Reality Check:\\Traditional, Generic Optimizers Suffice Across Batch Sizes}

\author{%
  Zachary Nado\thanks{Equal contribution}\\
  Google Research, Brain Team \\
  \texttt{znado@google.com} 
  \And
  Justin M. Gilmer\footnotemark[1]\\
  Google Research, Brain Team \\
  \texttt{gilmer@google.com} 
  \And
  Christopher J. Shallue\\
  Center for Astrophysics $|$\\Harvard \& Smithsonian,\\Cambridge, MA, USA \\
  \texttt{cshallue@cfa.harvard.edu} 
  \And
  Rohan Anil\\
  Google Research, Brain Team \\
  \texttt{rohananil@google.com} 
  \And
  George E. Dahl\\
  Google Research, Brain Team \\
  \texttt{gdahl@google.com} 
}

\begin{document}
\maketitle

\begin{abstract}
Recently the LARS and LAMB optimizers have been proposed for training neural networks faster using large batch sizes. LARS and LAMB add layer-wise normalization to the update rules of Heavy-ball momentum and Adam, respectively, and have become popular in prominent benchmarks and deep learning libraries. However, without fair comparisons to standard optimizers, it remains an open question whether LARS and LAMB have any benefit over traditional, generic algorithms. In this work we demonstrate that standard optimization algorithms such as Nesterov momentum and Adam can match or exceed the results of LARS and LAMB at large batch sizes. Our results establish new, stronger baselines for future comparisons at these batch sizes and shed light on the difficulties of comparing optimizers for neural network training more generally.
\end{abstract}

\newcommand{\srange}[2] {\lbrack\num{#1}, \num{#2}\rbrack}
\newcommand{\specialcell}[2][c]{%
  \begin{tabular}[#1]{@{}c@{}}#2\end{tabular}}

\section{Introduction}\label{sec:intro}
In recent years, hardware systems employing GPUs and TPUs have enabled neural network training programs to process dramatically more data in parallel than ever before.
The most popular way to exploit these systems is to increase the batch size in the optimization algorithm (i.e. the number of training examples processed per training step).
On many workloads, modern systems can scale to larger batch sizes without significantly increasing the time per step \citep{jouppi2017datacenter,wang2019benchmarking}, thus proportionally increasing the number of training examples processed per second.
If researchers can use this increased throughput to reduce the time required to train each neural network, then they should achieve better results by training larger models, using larger datasets, and by exploring new ideas more rapidly.

As the capacity for data parallelism continues to increase, practitioners can take their existing, well-tuned training configurations and re-train with larger batch sizes, hoping to achieve the same performance in less training time \citep[e.g.][]{ying2018image}.
On an idealized data-parallel system with negligible overhead from increasing the batch size, they might hope to achieve \textit{perfect scaling}, a proportional reduction in training time as the batch size increases.

However, achieving perfect scaling is not always straightforward.
Changing the batch size changes the training dynamics, requiring the training hyperparameters (e.g. learning rate) to be carefully re-tuned in order to maintain the same level of validation performance.\footnote{Although there are heuristics for adjusting the learning rate as the batch size changes, these heuristics inevitably break down sufficiently far from the initial batch size and it is also not clear how to apply them to other training hyperparameters (e.g. momentum).}
In addition, smaller batch sizes provide implicit regularization from gradient noise that may need to be replaced by other forms of regularization when the batch size is increased.
Finally, even with perfect tuning, increasing the batch size eventually produces diminishing returns.
After a critical batch size, the number of training steps cannot be decreased in proportion to the batch size -- the number of epochs must increase to match the validation performance of the smaller batch size.
See \citealt{shallue2019measuring} for a survey of the effects of data parallelism on neural network training. 
Once these effects are taken into account, there is no strong evidence that increasing the batch size degrades the maximum achievable performance on any workload.
At the same time, the ever-increasing capacity for data parallelism presents opportunities for new regularization techniques that can replace the gradient noise of smaller batch sizes and new optimization algorithms that can extend perfect scaling to larger batch sizes by using more sophisticated gradient information \citep{zhang2019algorithmic}.

\citet{you2017lars} proposed the LARS optimization algorithm in the hope of speeding up neural network training by exploiting larger batch sizes. LARS is a variant of stochastic gradient descent (SGD) with momentum \citep{polyak1964some} that applies layer-wise normalization before applying each gradient update.
Although it is difficult to draw strong conclusions from the results presented in the LARS paper,
\footnote{The modified AlexNet on ImageNet benchmark did not have well-established accuracy targets from prior work and LARS used a more general learning rate schedule than the momentum baseline. For ResNet-50 on ImageNet, LARS achieved sub-par accuracy numbers and was not compared to any other optimizer at the same batch size, leaving open the possibility that a generic optimizer would scale just as well as LARS.}
the MLPerf\footnote{MLPerf is a trademark of MLCommons.org.} Training benchmark\footnote{\url{https://mlperf.org/training-overview}} adopted LARS as one of two allowed algorithms in the closed division for ResNet-50 on ImageNet and it became the \textit{de facto} standard algorithm for that benchmark task.
With MLPerf entrants competing to find the fastest-training hyperparameters for LARS, the first place submissions in the two most recent MLPerf Training competitions used LARS to achieve record training speeds with batch sizes of 32,678 and 65,536, respectively.
No publications or competitive submissions to MLPerf have attempted to match these results with a standard optimizer (e.g. Momentum or Adam).
However, MLPerf entrants do not have a strong incentive (nor are necessarily permitted by the rules) to explore other algorithms because MLPerf Training is a systems benchmark that requires algorithmic equivalence between submissions to make fair comparisons.
Moreover, since the main justification for LARS is its excellent performance on ResNet-50 at large batch sizes, more work is needed to quantify any benefit of LARS over standard algorithms at any batch size.

\citet{you2019lamb} later proposed the LAMB optimizer to speed up pre-training for BERT \citep{devlin2018bert} using larger batch sizes after concluding that LARS was not effective across workloads.
LAMB is a variant of Adam \citep{kingma2014adam} that adds a similar layer-wise normalization step to LARS.
\citet{you2019lamb} used LAMB for BERT pre-training with batch sizes up to 65,536 and claimed that Adam cannot match the performance of LAMB beyond batch size 16,384.

In this paper, we demonstrate that standard optimizers, without any layer-wise normalization techniques, can match or improve upon the large batch size results used to justify LARS and LAMB. In Section~\ref{sec:resnet50}, we show that Nesterov momentum \citep{nesterov1983method} matches the performance of LARS on the ResNet-50 benchmark with batch size 32,768. We are the first to match this result with a standard optimizer.
In Section~\ref{sec:bert}, contradicting the claims in \citet{you2019lamb}, we show that Adam obtains better BERT pre-training results than LAMB at the largest batch sizes, resulting in better downstream performance metrics after fine-tuning.

In addition, we establish a new state-of-the-art for BERT pretraining speed, reaching an F1 score of 90.46 in 7,818 steps using Adam at batch size 65,536 (we report training speed in steps because our focus is algorithmic efficiency, but since we compare LARS and LAMB to simpler optimizers, fewer training steps corresponds to faster wall-time in an optimized implementation -- our BERT result with Adam also improves upon the wall-time record of LAMB reported in \citealt{you2019lamb}).
Taken together, our results establish stronger training speed baselines for these tasks and batch sizes, which we hope will assist future work aiming to accelerate training using larger batch sizes.

In addition to the contributions mentioned above, we demonstrate several key effects that are often overlooked by studies aiming to establish the superiority of new optimization algorithms.
We show that future work must carefully disentangle regularization and optimization effects when comparing a new optimizer to baselines.
We also report several under-documented details used to generate the best LARS and LAMB results, a reminder that future comparisons should document any novel tricks and include them in baselines.
Finally, our results add to existing evidence in the literature on the difficulty of performing independently rigorous hyperparameter tuning for optimizers and baselines. In particular, we show that the optimal shape of the learning rate schedule is optimizer-dependent (in addition to the scale), and that differences in the schedule can dominate optimizer comparisons at smaller step budgets and become less important at larger step budgets.

We made our code used for LARS experiments available at \url{https://github.com/google/init2winit}, and the official BERT codebase used for LAMB experiments can be found at \url{https://github.com/google-research/bert}.

\vspace{-1em}
\subsection{Related work}\label{sec:related-work}

\citet{shallue2019measuring} and \citet{zhang2019algorithmic} explored the effects of data parallelism on neural network training for different optimizers, finding no evidence that larger batch sizes degrade performance and demonstrating that different optimizers can achieve perfect scaling up to different critical batch sizes. \citet{you2017lars,you2019lamb} developed the LARS and LAMB optimizers  in the hope of speeding up training by achieving perfect scaling beyond standard optimizers. Many other recent papers have proposed new optimization algorithms for generic batch sizes or larger batch sizes \citep[see][]{schmidt2020descending}. \citet{choi2019empirical} and \citet{schmidt2020descending} demonstrated the difficulties with fairly comparing optimizers, showing that the hyperparameter tuning protocol is a key determinant of optimizer rankings. The MLPerf Training benchmark \citep{mattson2019mlperf} provides a competitive ranking of neural network training systems, but does not shed much light on the relative performance of optimizers because entrants are limited in the algorithms they can use and the hyperparameters they can tune. 

\vspace{-1em}
\section{Matching LARS on ImageNet}\label{sec:resnet50}
The MLPerf training benchmark for ResNet-50 v1.5 on ImageNet \citep{mattson2019mlperf} aims to reach 75.9\% validation accuracy in the shortest possible wall-clock time. In the closed division of the competition, entrants must choose between two optimizers, SGD with momentum or LARS, and are only allowed to tune a specified subset of the optimization hyperparameters, with the remaining hyperparameter values set by the competition rules.\footnote{\url{https://git.io/JtknD}}
The winning entries in the two most recent competitions used LARS with batch size 32,768 for 72 training epochs\footnote{\url{https://mlperf.org/training-results-0-6}} and LARS with batch size 65,536 for 88 training epochs,\footnote{\url{https://mlperf.org/training-results-0-7}} respectively.
\citet{kumar2019scale} later improved the training time for batch size 32,768 by reaching the target accuracy in 64 epochs.
These are currently the fastest published results on the ResNet-50 benchmark.
However, it has been unclear whether LARS was necessary to achieve these training speeds since no recent published results or competitive MLPerf submissions have used another optimizer.
In this section, we describe how we matched the 64 epoch, 32,768 batch size result of LARS using standard Nesterov momentum.\footnote{The 88 epoch, 65,536 batch size result is faster in terms of wall-clock time but requires more training epochs, indicating that it is beyond LARS's perfect scaling regime. Although LARS obtains diminishing returns when increasing the batch size from 32,768 to 65,536, future work could investigate whether Nesterov momentum drops off more or less rapidly than LARS.}

A fair benchmark of training algorithms or hardware systems must account for stochasticity in individual training runs. %
In the MLPerf competition, the benchmark metric is the mean wall-clock time of 5 trials after the fastest and slowest trials are excluded. Only 4 out of the 5 trials need to reach the target accuracy and there is no explicit limit on the number of times an entrant can try a different set of 5 trials.
Since our goal is to compare algorithms, rather than systems, we aim to match the LARS result in terms of training steps instead (but since Nesterov momentum is computationally simpler than LARS, this would also correspond to faster wall-clock time on an optimized system).
Specifically, we measure the median validation accuracy over 50 training runs with a fixed budget of 2,512 training steps\footnote{Corresponding to 64 training epochs in \citet{kumar2019scale}.} at a batch size of 32,768.
When we ran the published LARS training pipeline,\footnote{\url{https://git.io/JtsLQ}} LARS achieved a median accuracy of 75.97\% and reached the target in 35 out of 50 trials. We consider the LARS result to be matched by another optimizer if the median over 50 trials exceeds the target of 75.9\%.%

\vspace{-0.5em}
\subsection{Nesterov momentum at batch size 32k}

This section describes how we used the standard Nesterov momentum optimizer to train the ResNet-50 v1.5 on ImageNet to 75.9\% validation accuracy in 2,512 update steps at a batch size of 32,768, matching the best published LARS result at this batch size. Although we implemented our own training program, the only logical changes we made to the published LARS pipeline were to the optimizer and the optimization hyperparameters. Our model implementation and data pre-processing pipeline were identical to those required under the MLPerf closed division rules (see Appendix~\ref{appendix:experiment-details}).

We present two Nesterov momentum hyperparameter configurations that achieve comparable performance to LARS. Configuration A achieved a median accuracy of 75.97\% (the same as LARS) and reached the target accuracy in 34 out of 50 trials. Configuration B is a modified version of Configuration A designed to make as few changes as possible to the LARS hyperparameters; it achieved a median accuracy of 75.92\% and reached the target in 29 out of 50 trials. 
See Appendix~\ref{appendix:nesterov_config} for the complete hyperparameter configurations.

To achieve these results, we tuned the hyperparameters of the training pipeline from scratch using Nesterov momentum. We ran a series of experiments, each of which searched over a hand-designed hyperparameter search space using quasi-random search \citep{bousquet2017critical}. Between each experiment, we modified the previous search space and/or tweaked the training program to include optimization tricks and non-default hyperparameter values we discovered in the state-of-the-art LARS pipeline. The full sequence of experiments we ran, including the number of trials, hyperparameters tuned, and search space ranges, are provided in Appendix~\ref{appendix:historical_search_spaces_nesterov_resnet50}. Once we had matched the LARS result with Configuration A, we tried setting each hyperparameter to its value in the LARS pipeline in order to find the minimal set of changes that still achieved the target result, producing Configuration B. The remainder of this section describes the hyperparameters we tuned and the techniques we applied on the journey to these results.

\vspace{-0.5em}
\subsubsection{Nesterov Momentum Optimizer}\label{sec:nesterov-momentum}

Nesterov momentum is a variant of classical or ``heavy-ball'' momentum defined by the update rule
\newcommand{\loss}{\ell}
\newcommand{\grad}{\nabla}
\begin{equation*}
\begin{aligned}
    &v_{t+1} = \mu v_{t} + \grad \loss(\theta_t), \\
    &\theta_{t+1} = \theta_{t} - \eta_t \left( \mu v_{t+1} + \grad \loss(\theta_{t}) \right),
\end{aligned}
\end{equation*}
where $v_0 = 0$, $\theta_t$ is the vector of model parameters after $t$ steps, $\grad \loss(\theta_t)$ is the gradient of the loss function $\ell(\theta)$ averaged over a batch of training examples, $\mu$ is the momentum, and $\eta_t$ is the learning rate for step $t$. We prefer Nesterov momentum over classical momentum because it tolerates larger values of its momentum parameter \citep{sutskever2013importance} and sometimes outperforms classical momentum, although the two algorithms perform similarly on many tasks \citep{shallue2019measuring,choi2019empirical}.
We tuned the Nesterov momentum $\mu$ in Configurations A and B. We discuss the learning rate schedule $\{ \eta_t \}$ separately in Section~\ref{sec:learning-rate}.

\vspace{-0.5em}
\subsubsection{Batch normalization}\label{sec:batch-norm}

The ResNet-50 v1.5 model uses batch normalization \citep{ioffe2015batch}, defined as
\begin{equation*}
    \texttt{BN}(x^{(l)}) = \left( \frac{x^{(l)} - \texttt{mean}(x^{(l)})}{\sqrt{\texttt{var}(x^{(l)}) + \epsilon}} \right) \times \gamma^{(l)} + \beta^{(l)},
\end{equation*}
where $x^{(l)}$ is a vector of pre-normalization outputs from layer $l$, $\texttt{mean}(\cdot)$ and $\texttt{var}(\cdot)$ denote the element-wise sample mean and variance across the batch of training examples,\footnote{In a distributed training environment the mean and variance are commonly computed over a subset of the full batch. The LARS pipeline uses a ``virtual batch size'' of 64, which we also use to avoid changing the training objective \citep{hoffer2017train}.} and $\gamma^{(l)}$ and $\beta^{(l)}$ are trainable model parameters.

Batch normalization introduces the following tuneable hyperparameters: $\epsilon$, the small constant added to the sample variance; the initial values of $\gamma^{(l)}$ and $\beta^{(l)}$; and $\rho$, which governs the exponential moving averages of the scaling factors used in evaluation. The LARS pipeline uses $\epsilon = \num{e-5}$ and $\rho = 0.9$. It sets the initial value of $\beta^{(l)}$ to 0.0 everywhere, but the initial value of $\gamma^{(l)}$ depends on the layer: it sets $\gamma^{(l)}$ to 0.0 in the final batch normalization layer of each residual block, and to 1.0 everywhere else. In Configuration A, we tuned $\epsilon$, $\rho$, and $\gamma_0$, the initial value of $\gamma^{(l)}$ in the final batch normalization layer of each residual block. In Configuration B, we used the same values as LARS for $\epsilon$ and $\rho$, but we found that choosing $\gamma_0$ between 0.0 and 1.0 was important for matching the LARS result with Nesterov momentum.

\subsubsection{Regularization}\label{sec:regularization}

In Configuration A, we tuned both the L2 regularization coefficient $\lambda$ and label smoothing coefficient $\tau$ \citep{szegedy2016rethinking}. The LARS pipeline uses $\lambda = \num{e-4}$ and $\tau = 0.1$.
Crucially, the LARS pipeline does not apply L2 regularization to the bias variables of the ResNet model nor the batch normalization parameters $\gamma^{(l)}$ and $\beta^{(l)}$ (indeed, the published LARS pipeline does not even apply LARS to these parameters -- it uses Heavy-ball momentum). This detail is extremely important for both LARS and Nesterov momentum to achieve the fastest training speed. Configuration B used the same $\lambda$ and $\tau$ as Configuration A.
\begin{wraptable}[12]{r}{0.35\linewidth}
\centering
\setlength{\extrarowheight}{3.5pt}
\begin{tabular}{|c|c|c|}
\hline
 & Nesterov & LARS \\ \hline
$p_\text{warmup}$ & 2 & 1 \\ \hline
$\eta_\text{peak}$ & 7.05 & 29.0 \\ \hline
$\eta_\text{final}$ & $\num{6e-6}$ & $\num{e-4}$ \\ \hline
$1 - \mu$ & 0.02397 & 0.071 \\ \hline
$\lambda$ & $\num{5.8e-5}$ & $\num{e-4}$ \\ \hline
$\tau$ & 0.15 & 0.10 \\ \hline
$\gamma_0$ & 0.4138 & 0.0 \\ \hline
\end{tabular}
\caption{The hyperparameters of Configuration B that differ from state-of-the-art LARS at batch size 32,768 \citep{kumar2019scale}.}\label{table:best_nesterov_lars_hparams_resnet50}
\end{wraptable}
\subsubsection{Learning rate schedule}\label{sec:learning-rate}
The LARS pipeline uses a piecewise polynomial schedule
\begin{equation*}
    \eta_t = 
    \begin{cases}
      \eta_\text{init} + (\eta_\text{peak} - \eta_\text{init}) \left( \frac{t}{t_\text{warmup}} \right)^{p_\text{warmup}}, & t \le t_\text{warmup} \\
      \eta_\text{final} + \left(\eta_\text{peak} - \eta_\text{final}\right) \left(\frac{T - t}{T - t_\text{warmup}}\right) ^{p_\text{decay}} & t > t_\text{warmup},
    \end{cases}
\end{equation*}
with $\eta_\text{init} = 0.0$, $\eta_\text{peak} = 29.0$, $\eta_\text{final} = \num{e-4}$, $p_\text{warmup} = 1$, $p_\text{decay} = 2$, and $t_\text{warmup} =706$ steps. In Configuration A, we re-tuned all of these hyperparameters with Nesterov momentum. In Configuration B, we set $\eta_\text{init}$, $p_\text{decay}$, and $t_\text{warmup}$ to the same values as LARS, changing only $p_\text{warmup}$ from 1 to 2 and re-scaling $\eta_\text{peak}$ and $\eta_\text{final}$.

\vspace{-0.5em}
\subsubsection{Comparing Nesterov momentum and LARS}\label{sec:nesterov-vs-lars}
Table~\ref{table:best_nesterov_lars_hparams_resnet50} shows the hyperparameter values for Configuration B that differ from the state-of-the-art LARS pipeline. Aside from re-tuning the momentum, learning rate scale, and regularization hyperparameters (whose optimal values are all expected to change with the optimizer), the only changes are setting $p_\text{warmup}$ to 2 instead of 1 and re-tuning $\gamma_0$.
\begin{wrapfigure}[14]{r}{0.4\linewidth}
    \centering
    \includegraphics[width=\linewidth]{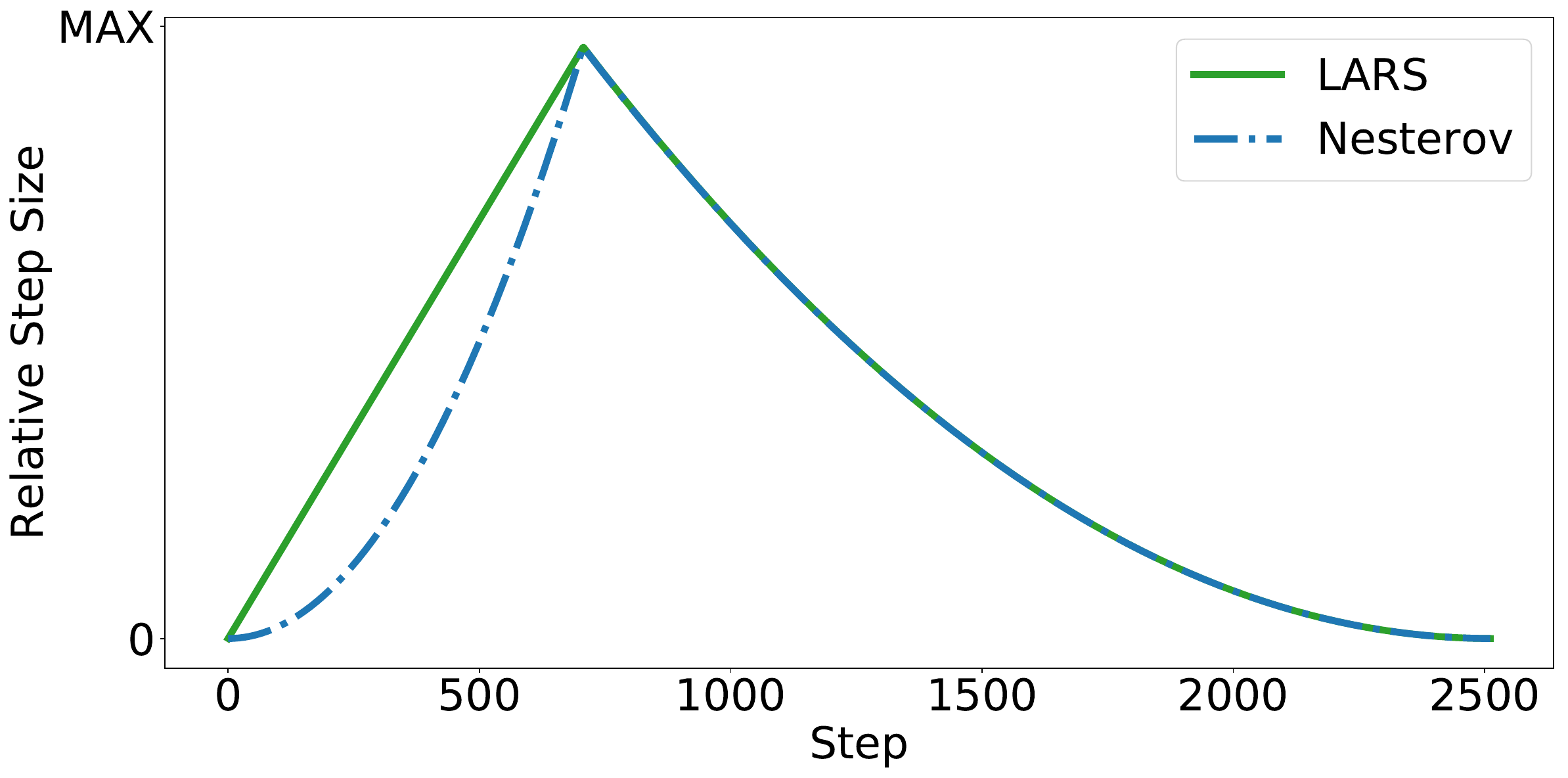}
    \caption{The learning rate schedules of LARS and Nesterov momentum Configuration B. Aside from re-scaling, the only difference is setting the warmup polynomial power to 2 instead of 1.}
    \label{fig:mlperf_lr_schedules}
\end{wrapfigure}

Figure~\ref{fig:mlperf_lr_schedules} shows the LARS learning rate schedule compared to the Nesterov momentum schedule. Even though these schedules are similar, we found that each optimizer had a different optimal value of the warmup polynomial power.
As Table~\ref{table:generalization} shows, Nesterov momentum performs better with $p_\text{warmup} = 2$ instead of 1, while the opposite is true with LARS.
As discussed in \citet{agarwal2020disentangling}, optimizers can induce implicit step size schedules that strongly influence their training dynamics and solution quality, and it appears from Table~\ref{table:generalization} that the implicit step sizes of Nesterov momentum and LARS may evolve differently, causing the shapes of their optimal learning rate schedules to differ.

Although the main concern of a practitioner is validation performance, the primary task of an optimization algorithm is to minimize training loss. Table~\ref{table:generalization} shows that Nesterov momentum achieves higher training accuracy than LARS, despite similar validation performance. Thus, it may be more appropriate to consider the layerwise normalization of LARS to be a regularization technique, rather than an optimization technique.

Spending even more effort tuning LARS or Nesterov momentum would likely further improve the current state-of-the-art for that optimizer. Meaningful optimizer comparisons are only possible with independent and equally intensive tuning efforts, and we do not claim that either optimizer outperforms the other on this benchmark.
That said, if the main evidence for LARS's utility as a ``large-batch optimizer'' is its performance on this particular benchmark, then more evidence is needed to quantify any benefit it has over traditional, generic optimizers like Nesterov momentum. 
\vspace{-0.3em}
\begin{table}[h]
\centering
\setlength{\extrarowheight}{3.5pt}
\begin{tabular}{|c|c|c|}
\hline
$p_\text{warmup}$ & Nesterov & LARS \\ \hline
1 &  75.79\% & 75.97\% \\ \hline
2 & 75.92\% & 75.69\% \\ \hline
\end{tabular}
\qquad
\begin{tabular}{|c|c|c|}
\hline
Optimizer & Train Acc & Test Acc \\ \hline
Nesterov & $78.97 \%$ & $75.93 \%$ \\ \hline
LARS & $78.07 \%$ & $75.97 \%$ \\ \hline
\end{tabular}
\caption{\textbf{(Left)} The best warmup schedule differs for Nesterov momentum and LARS. Values are medians over 50 training runs after setting $p_\text{warmup}$ without retuning other hyperparameters. \textbf{(Right)} Median train and test accuracies over 50 training runs for Nesterov momentum Configuration B and LARS.}\label{table:generalization}
\end{table}

\subsection{Lessons learned}
In hindsight, it was only necessary to make a few changes to the LARS pipeline to match its performance at batch size 32,768 with Nesterov momentum. However, Table~\ref{fig:mlperf_lr_schedules} does not accurately represent the effort required when attempting to match a highly tuned training-speed benchmark.

Firstly, as described in Sections~\ref{sec:batch-norm} and ~\ref{sec:regularization}, the strong results of LARS depend partly on a few subtle optimization tricks and non-default values of uncommonly-tuned hyperparameters. Fortunately, in this case we could discover these tricks by examining the open-source code required for MLPerf submissions, but machine learning research papers do not always report these important details. Researchers can easily waste a lot of experiments and produce misleading results before getting all of these details right.
We demonstrate the importance of adding these tricks to our Nesterov momentum pipeline in Appendix~\ref{appendix:nesterov-ablations}; without these tricks (or some new tricks), we likely would not have been able to match the LARS performance.

Secondly, the learning rate schedule really matters when trying to maximize performance with a relatively small step budget. Both LARS and Nesterov momentum are sensitive to small deviations from the optimized learning rate schedules in Figure~\ref{fig:mlperf_lr_schedules}, and neither schedule works as well for the other optimizer. Although relatively minor changes were sufficient to match LARS with Nesterov momentum, there is no way to know \textit{a priori} how the optimal schedule will look for a new optimizer \cite{wu2018understanding}. Even in toy settings where the optimal learning rate schedule can be derived, it does not fit into commonly used schedule families and depends strongly on the optimizer \cite{zhang2019algorithmic}. 
Indeed, this problem applies to the other optimization hyperparameters as well: it is extremely difficult to know which are worth considering ahead of time.
Finally, even when we narrowed down our hyperparemeter search spaces around the optimal point, the volume of our search spaces corresponding to near-peak performance was small, likely due to the small step budget \citep{shallue2019measuring}. We investigate how these effects change with a less stringent step budget in Section~\ref{sec:step_budget}.

\vspace{-1em}
\section{Stronger BERT pretraining speed baselines}\label{sec:bert}
\citet{you2019lamb} developed the LAMB optimizer in the hope of speeding up training for BERT-Large \citep[Bidirectional Encoder Representations from Transformers,][]{devlin2018bert}.
BERT training consists of two phases. The ``pretraining'' phase has two objectives: (1) predicting masked tokens based on the rest of the sequence (a masked language model), and (2) predicting whether two given sentences follow one from another. Finally, the ``fine-tuning'' phase refines the model for a downstream task of interest.  BERT pretraining takes a considerable amount of time (up to 3 days on 16 Cloud TPU-v3 chips \cite{jouppi2017datacenter}), whereas the fine-tuning phase is typically much faster. Model quality is typically assessed on the downstream metrics, not on pretraining loss, making BERT training a somewhat awkward benchmark for optimization research.

\citet{you2019lamb} used LAMB for BERT pretraining with batch sizes up to 65,536 and claimed that LAMB outperforms Adam batch size 16,384 and beyond. The LAMB optimizer has since appeared in several NLP toolkits, including as Microsoft DeepSpeed and NVIDIA Multi-node BERT training, and as a benchmark task in MLPerf v0.7.\footnote{We do not consider the MLPerf task in this paper since it is a warm-start, partial training task.} 

As shown in Table~\ref{table:bert-results}, we trained Adam (with decoupled weight decay) baselines that achieve better results than both the LAMB and Adam results reported in \citet{you2019lamb}. Our new Adam baselines obtain better F1 scores on the development set of the SQuaD v1.1 task in the same number of training steps as LAMB for both batch size 32,768 and the hybrid 65,536-then-32,768 batch size training regime in \citet{you2019lamb}. We also ran Adam at batch size 65,536 to reach nearly the same F1 score as the hybrid batch size LAMB result, but in much fewer training steps. We believe 7,818 steps is a new state-of-the-art for BERT pretraining speed \citep[in our experiments, it also improves upon the 76-minute record claimed in][]{you2019lamb}. Additionally, at batch size  32,768 our Adam baseline got a better pretraining loss of 1.277 compared to LAMB's 1.342.
\begin{wraptable}[9]{r}{0.5\linewidth}
\centering
 \begin{tabular}{|c|c|c|c|}
\hline
Batch size & Step budget & LAMB & Adam \\
\hline
32k & 15,625 & 91.48 & \textbf{91.58} \\
\hline
65k/32k & 8,599 & 90.58 & \textbf{91.04} \\
\hline
65k & 7,818 & -- & \textbf{90.46} \\
\hline
\end{tabular}
\caption{Using Adam for pretraining exceeds the reported performance of LAMB in \citet{you2019lamb} in terms of F1 score on the downstream SQuaD v1.1 task.}
\label{table:bert-results}
\end{wraptable}

We used the same experimental setup as \citet{you2019lamb}, including two pretraining phases with max sequence lengths of 128 and then 512.
In order to match \citet{you2019lamb}, we reported the F1 score on the downstream SQuaD v1.1 task as the target metric, although this metric introduces potential confounds: optimization efficiency should be measured on the training task using training and held-out data sets. Fortunately, in this case better pretraining performance correlated a with higher F1 score after fine-tuning. See Appendix~\ref{appendix:bert_experiment_details} for additional experiment details. We tuned Adam hyperparameters independently for each pretraining phase, specifically learning rate $\eta$, $\beta_1$, $\beta_2$, the polynomial power for the learning rate warmup $p_{warmup}$, and weight decay $\lambda$, using quasi-random search \citep{bousquet2017critical}. See Appendix~\ref{appendix:adam_bert_hparams} for the search spaces.

In addition to hyperparmeter tuning, our improved Adam results at these batch sizes are also likely due to two implementation differences. First, the Adam implementation in \citet{you2019lamb} comes from the BERT open source code base, in which Adam is missing the standard bias correction.\footnote{\url{https://git.io/JtY8d}} The Adam bias correction acts as an additional step size warm-up, thereby potentially improving the stability in the initial steps of training. Second, the BERT learning rate schedule had a discontinuity at the start of the decay phase due to the learning rate decay being incorrectly applied during warm-up \footnote{See \url{https://git.io/JtnQW} and \url{https://git.io/JtnQ8}.} (see Figure~\ref{fig:mlperf_bert_schedules} in Appendix~\ref{appendix:experiment-details}). This peculiarity is part of the official BERT release and is present in 3000+ copies of the BERT Training code on GitHub.

\vspace{-1em}
\section{Investigating a less stringent step budget}\label{sec:step_budget}
Part of what makes comparing optimizers so difficult is that the hyperparameter tuning tends to dominate the comparisons \citep{choi2019empirical}. Moreover, tuning becomes especially difficult when we demand a fixed epoch budget even when dramatically increasing the batch size \citep{shallue2019measuring}.
Fixing the epoch budget as the batch size increases is equivalent to demanding perfect scaling (i.e. that the number of training steps decreases by the same factor that the batch size is increased).
We can view the role of hyperparameter tuning for large batch training as resisting the inevitable end of perfect scaling.
For example, it might be possible to extend perfect scaling using delicately tuned learning rate schedules, but comparing optimizers under these conditions can make the learning rate schedule dominate the comparison by favoring some algorithms over others.
Therefore, in order to better understand the behavior of LARS and LAMB compared to Nesterov Momentum and Adam, we ran additional ResNet-50 experiments with a more generous 6,000 step budget (vs 2,512 in Section~\ref{sec:resnet50}) and a more simplistic cosine learning rate schedule. At batch size 32,768, this budget should let us reach better validation accuracy than the MLPerf target of 75.9\%.

Although not mentioned in \citet{you2017lars}, the state-of-the-art MLPerf pipeline for ``LARS'' actually uses both LARS and Heavy-ball Momentum, with Momentum applied to the batch normalization and ResNet bias parameters and LARS applied to the other parameters. \citet{you2019lamb} does not mention whether LAMB was only applied to some parameters and not others.
If layerwise normalization can be harmful for some model parameters, this is critical information for practitioners using LARS or LAMB, since it might not be obvious which optimizer to apply to which parameters.
To investigate this, we trained both pure LARS and LAMB configurations, as well as configurations that did not apply layerwise normalization to the batch normalization and ResNet bias parameters.
Moreover, LAMB's underlying Adam implementation defaults to $\epsilon=10^{-6}$, rather than the typical $10^{-7}$ or $10^{-8}$. In some cases, $\epsilon$ can be a critical hyperparameter for Adam \citep{choi2019empirical}, so we included Adam configurations with both $\epsilon = 10^{-6}$ and $\epsilon = 10^{-8}$.

Table~\ref{table:cosine_schedule} shows the validation accuracy of these different configurations after training for 6,000 steps with batch size 32,768. In every case, we used a simple cosine decay learning rate schedule and tuned the initial learning rate and weight decay using quasi-random search. We used momentum parameters of 0.98 for Nesterov momentum and 0.929 for LARS, respectively, based on the tuned values from Section~\ref{sec:resnet50}. We used default hyperparameters for Adam and LAMB except where specified. We set all other hyperparameters to the same values as the state-of-the-art LARS pipeline, except we set $\gamma_0 = 1.0$. See Appendix~\ref{appendix:cosine_details} for more details. As expected, highly tuned learning rate schedules and optimizer hyperparameters are no longer necessary with a less stringent step budget. Multiple optimizer configurations in Table~\ref{table:cosine_schedule} exceed the MLPerf target accuracy of 75.9\% at batch size 32,768 with minimal tuning. Training with larger batch sizes is \emph{not} fundamentally unstable: stringent step budgets make hyperparameter tuning trickier.
\begin{wraptable}[22]{l}{0.5\linewidth}
\centering
\setlength{\extrarowheight}{3.5pt}
\begin{tabular}{|c|c|c|}
\hline
\specialcell{Weights\\Optimizer} & \specialcell{Bias/BN\\Optimizer} & Top-1 \\ \hline
Nesterov & Nesterov & 76.7 \\ \hline %
LARS & Momentum & 76.9 \\ \hline %
LARS & LARS & 76.9 \\ \hline %
Adam ($\epsilon=10^{-8}$) & Adam ($\epsilon=10^{-8}$) & 76.2 \\ \hline %
Adam ($\epsilon=10^{-6}$) & Adam ($\epsilon=10^{-6}$) & 76.4 \\ \hline %
LAMB & LAMB & 27.3 \\ \hline %
LAMB & Adam ($\epsilon=10^{-8}$) & 76.3 \\ \hline %
LAMB & Adam ($\epsilon=10^{-6}$) & 76.3 \\ \hline %
\end{tabular}
\caption{Validation accuracy of ResNet-50 on ImageNet trained for 6,000 steps instead of 2,512. The second column is the optimizer that was applied to the batch norm and ResNet bias variables. We report the median top-1 accuracy over 5 seeds of the best hyperparameter setting in a refined search space. See Appendix~\ref{appendix:cosine_details} for details.}
\label{table:cosine_schedule}
\end{wraptable}

In Table~\ref{table:cosine_schedule}, ``pure LAMB'' performs extremely poorly: LAMB only obtains reasonable results when it is \textit{not} used on the batch normalization and ResNet bias parameters, suggesting that layerwise normalization can indeed be harmful on some parameters. ``Pure LARS'' and Nesterov momentum perform roughly the same at this step budget, but the MLPerf LARS pipeline, which is tuned for a more stringent step budget, does not use LARS on all parameters, at least suggesting that the optimal choice could be budget-dependent.

Many new neural net optimizers, including LAMB, are introduced alongside claims that the new optimizer does not require any---or at least minimal---tuning. Unfortunately, these claims require a lot of work to support, since they require trying the optimizer on new problems without using those problems during the development of the algorithm. Although our experiments here are not sufficient to determine which optimizers are easiest to tune, experiments like these that operate outside the regime of highly tuned learning rate schedules can serve as a starting point.
In this experiment, LARS and LAMB do not appear to have an advantage in how easy they are to tune even on a dataset and model that were used in the development of both of those algorithms. LAMB is a variant of Adam and performs about the same as Adam with the same value of $\epsilon$; LARS is more analogous to Momentum and indeed Nesterov momentum and LARS have similar performance.

\vspace{-1em}
\section{Discussion}\label{sec:discussion}
\vspace{-0.5em}
Our results show that standard, generic optimizers suffice for achieving strong results across batch sizes.
Therefore, any research program to create new optimizers for training at larger batch sizes must start from the fact that Momentum, Adam, and likely other standard methods work fine at batch sizes as large as those considered in this paper. The LARS and LAMB update rules have no more to do with the batch size (or ``large" batches) than the Momentum or Adam update rules.
Although \citet{you2019lamb} presented convergence rate bounds for LARS and LAMB to support their claims of superior performance, we show in Appendix~\ref{appendix:converge-proofs} that Adam satisfies a similar bound to LAMB. These bounds all rely on very unrealistic assumptions.\footnote{All convergence bounds assume no momentum is used, and the $L_{avg}$ bound for LAMB also assumes $\beta_2 = 0$, when it is typically 0.999. Additionally, $L_{avg}$ could still be large if $L_\infty$ is large, but we leave an empirical analysis of this to future work.} Most of all, they are loose upper bounds on the worst case behavior of the algorithms, not accurate reflections of optimizer performance in reality.
Whether layer-wise normalization can be useful for optimization or regularization remains an open question. However, if LARS and LAMB have any advantage over standard techniques, it is not that they work dramatically better on the tasks and batch sizes in \citet{you2017lars,you2019lamb}.
This is not to suggest that there is nothing interesting about studying neural network optimization at larger batch sizes. For example, as gradient noise decreases, there may be opportunities to harness curvature information and extend the region of perfect scaling \citep{zhang2019algorithmic}. However, there is currently no evidence that LARS and LAMB scale better than Momentum and Adam.

Our primary concern in this paper has been matching the state of the art---and establishing new baselines---for \emph{training speed} measurements of the sort used to justify new techniques and algorithms for training with larger batch sizes. In contrast, many practitioners are more concerned with obtaining the best possible validation error with a somewhat flexible training time budget. Part of the reason why matching LARS at batch size 32,768 was non-trivial is because getting state of the art training speed requires several tricks and implementation details that are not often discussed. It was not obvious to us \emph{a priori} which ones would prove crucial. These details do not involve changes to the optimizer, but they interact with the optimizer in a regime where all hyperparameters need to be well tuned to stay competitive, making it necessary to re-tune everything for a new optimizer.

In neural network optimization research, training loss is rarely discussed in detail and evaluation centers on validation/test performance since that is what practitioners care most about. However, although we shouldn't \emph{only} consider training loss, it is counter-intuitive and counter-productive to elide a careful investigation of the actual objective of the optimizer. If a new optimizer achieves better test performance, but shows no speedup on training loss, then perhaps it is \emph{not} a better optimizer so much as an indirect regularizer.
\footnote{Deep learning folk wisdom is that ``any method to make training less effective can serve as a regularizer," whether it is a bug in gradients or a clever algorithm.}
Indeed, in our experiments we found that Nesterov momentum achieves noticeably better training accuracy on ResNet-50 than the LARS configuration we used, despite reaching roughly the same validation accuracy. Properly disentangling possible regularization benefits from optimization speed-ups is crucial if we are to understand neural network training, especially at larger batch sizes where we lose some of the regularization effect of gradient noise. Hypothetically, if the primary benefit of a training procedure is regularization, then it would be better to compare the method with other regularization baselines than other optimizers.

Ultimately, we only care about batch size to the extent that higher degrees of data parallelism lead to faster training. Training with a larger batch size is a means, not the end goal.
New optimizers---whether designed for generic batch sizes or larger batch sizes---have the potential to dramatically improve algorithmic efficiency across multiple workloads, but our results show that standard optimizers can match the performance of newer alternatives on the workloads we considered.
Indeed, despite the legion of new update rule variants being proposed in the literature, standard Adam and Momentum remain the workhorses of practitioners and researchers alike, while independent empirical comparisons consistently find no clear winner when optimizers are compared across a variety of workloads \citep{schmidt2020descending}. Meanwhile, as \citet{choi2019empirical} and our results underscore, comparisons between optimizers crucially depend on the effort spent tuning hyperparameters for each optimizer. Given these facts, we should regard with extreme caution studies claiming to show the superiority of one particular optimizer over others. Part of the issue stems from current incentives in the research community; we overvalue the novelty of new methods and undervalue establishing strong baselines to measure progress against. This is particularly problematic in the study of optimizers, where the learning rate schedule is arguably more important than the choice of the optimizer update rule itself! As our results show, the best learning rate schedule is tightly coupled with the optimizer, meaning that tuning the learning rate schedule for a new optimizer will generally favor the new optimizer over a baseline unless the schedule of the baseline is afforded the same tuning effort.

\vspace{-1em}
\section{Conclusion}\label{sec:conclusion}
In this work, we demonstrated that standard optimizers, without any layer-wise normalization techniques, can match or exceed the large batch size results used to justify LARS and LAMB. 
Future work attempting to argue that a new algorithm is useful by comparing to baseline methods or results, including those established in this paper, faces a key challenge in showing that the gains are due to the new method and not merely due to better tuning or changes to the training pipeline (e.g. regularization tricks).
Although gains from tuning will eventually saturate, we can, in principle, always invest more effort in tuning and potentially get better results for any optimizer. However, our goal should be developing optimizers that work better across many different workloads when taking into account the amount of additional tuning they require.

Moving forward, if we are to reliably make progress we need to rethink how we compare and evaluate new optimizers for neural network training. Given how sensitive optimizer performance is to the hyperparameter tuning protocol and how difficult it is to quantify hyperparameter tuning effort, we can't expect experiments with self-reported baselines to always lead to fair comparisons. Ideally, new training methods would be evaluated in a standardized competitive benchmark, where submitters of new optimizers do not have full knowledge of the evaluation workloads. Some efforts in this direction have started, for instance the MLCommons Algorithmic Efficiency Working Group\footnote{\url{https://mlcommons.org/en/groups/research-algorithms/}}, but more work needs to be done to produce incentives for the community to publish well-tuned baselines and to reward researchers that conduct the most rigorous empirical comparisons.

\bibliography{main}
\bibliographystyle{plainnat}

\clearpage

\appendix
\section{Convergence Proofs}\label{appendix:converge-proofs}
To support our larger point about the irrelevance of this type of result when comparing LAMB and Adam, below we derive a convergence bound for Adam in a similar manner to the LAMB bound in \citet{you2019lamb}. Note that all of these bounds are loose upper bounds on the worst case behavior of the algorithms, so there is no reason that comparing them reflects the relative behaviors of optimizers in reality. For example in Equation~\ref{eq:adam-eq-4} below, we follow similar operations as the LAMB bound derivation and simply switch a $-$ to a $+$ for algebraic convenience.

We define the following as our optimization objective
\begin{align}
\label{eq:proof_obj}
\min_{x \in \mathbb{R}^d} f(x) := \mathbb{E}_{s \in \mathbb{P}}[\ell(x, s)] + \frac{\lambda}{2} \|x\|^2,
\end{align}
with an optimal solution(s) $x^*$. $x \in \mathbb{R}^d$ are the neural network parameters, $\ell$ a smooth and possibly nonconvex loss function, $\mathbb{P}$ a data distribution, and $\lambda$ the regularization strength.

Let $T$ be the number of training steps, $h$ the number of neural network layers, $b$ the batch size, $\eta$ the learning rate, $n$ the mini-batch size, and $\phi(v):\mathbb{R}^{+} \rightarrow \mathbb{R}^{+}$ a function that is layerwise multiplied by the learning rate in LARS and LAMB updates.
Let $L$ be a vector of the layerwise Lipschitz constants for the neural network, and $L_{avg}$ the mean of $L$.
Let $s$ be a training step uniformly sampled from $\{1, 2, ..., T\}$.

We define the stochastic minibatch estimate of the true gradient as $\mathbb{E}[g^{(i)}] = \nabla_i f(x)$ and assume that its variance is bounded by $\mathbb{E}\left[g^{(i)} - \nabla_i f(x)\right]^2 \leq \sigma_i^2$ layerwise for a vector of standard deviations $\sigma := \left[\sigma^{(1)}, \ldots \sigma^{(h)}\right]$ and elementwise for $\tilde{\sigma} := \left[\tilde{\sigma}^{(1)}, \ldots \tilde{\sigma}^{(h)}\right]$.

Next, let $\eta_t = \eta = \sqrt{\tfrac{2(f(x_1) - f(x^*))}{\alpha_u^2 \|L\|_1 T}}$ $\forall t \in [T]$, $b=T$, $\alpha_l \leq \phi(v) \leq \alpha_u$  $\forall v > 0$, $\alpha_l, \alpha_u > 0$. Crucially, additionally let $b = T, \beta_1 = 0, \lambda = 0$. Under these conditions \citet{you2019lamb} show the convergence rate for LARS is
\begin{align*}
\left(\mathbb{E}\left[\frac{1}{\sqrt{h}}\sum_{i=1}^h \|\nabla_i f(x_s)\|\right]\right)^2 \leq \mathcal{O}\left(\frac{(f(x_1) - f(x^*))L_{avg}}{T} + \frac{\|\sigma \|^2_1}{Th}\right).
\end{align*}
They also derive the convergence rate of LAMB as
\begin{align*}
 \mathbb{E}[\|\nabla f(x_a)\|^2] &\leq \mathcal{O}\left(\sqrt{\frac{G^2 d}{h(1 - \beta_2)}} \times \left[\sqrt{\frac{2(f(x_1) - f(x^*))\|L\|_1}{T}} + \frac{\|\tilde{\sigma}\|_1}{\sqrt{T}} \right]\right).
\end{align*}
Additionally, for $\beta_2 = 0$, the convergence rate of LAMB can be derived as
\begin{align*}
\left(\mathbb{E}\left[\frac{1}{\sqrt{d}}\|\nabla f(x_a)\|_1\right]\right)^2 &\leq  \mathcal{O}\left(\frac{(f(x_1) - f(x^*))L_{avg}}{T} + \frac{\|\tilde{\sigma}\|^2_1}{Th}\right),
\end{align*}

Below we derive a similar bound for the $\beta_2 > 0$ case for Adam updates. We note that the $\beta_2 = 0$ case where the bound depends on $L_{avg}$ instead of $||L||_1$ can be very similarly derived for Adam, but is also a very unrealistic condition in practice.

\begin{proof}
Under the assumption $\beta_1=0, \lambda=0$, one could write the Adam update rule as follows:

$$
x_{t+1}^{(i)}  = x_{t}^{(i)} - \eta_t \sqrt{1 -\beta_2^t} \frac{g_{t}^{(i)}}{\sqrt{v_{t}^{(i)}}}, 
$$
where $v_{t} = \beta_{2} v_{t-1} + (1 - \beta_{2}) g_{t}^2$ for all $i \in [h]$.

Since the function $f$ is $L$-smooth, we have the following:
\begin{align}
f(x_{t+1}) &\leq f(x_t) + \langle \nabla_i f(x_t), x_{t+1}^{(i)} - x_{t}^{(i)} \rangle + \sum_{i=1}^h \frac{L_i}{2} \|x_{t+1}^{(i)} - x_t^{(i)}\|^2 \nonumber \\
&\leq f(x_t) \underbrace{- \eta_t  \sum_{i=1}^h \sum_{j=1}^{d_i} [\nabla_i f(x_t)]_j  \sqrt{1 -\beta_2^t} \frac{g_{t, j}^{(i)}}{\sqrt{v_{t, j}^{(i)}}}}_{T_1} + \sum_{i=1}^h \frac{L_i\eta_t^2d}{2(1-\beta_2)}
\label{eq:adam-eq-1}
\end{align}

Where the last term comes from the fact that $1 - \beta_2^t \leq 1$. We bound term $T_1$ in the following manner, in line with \citep{you2019lamb}:

\begin{align}
T_1  &= - \eta_t  \sum_{i=1}^h \sum_{j=1}^{d_i} [\nabla_i f(x_t)]_j  \sqrt{1 -\beta_2^t} \frac{g_{t, j}^{(i)}}{\sqrt{v_{t, j}^{(i)}}} \nonumber \\
&\leq - \eta_t \sum_{i=1}^h \sum_{j=1}^{d_i} \frac{\sqrt{1 - \beta_2}}{G}  [\nabla_i f(x_t)]_j g_{t, j}^{(i)} \nonumber \\
&\qquad \qquad - \eta_t \sum_{i=1}^h \sum_{j=1}^{d_i} \left([\nabla_i f(x_t)]_j  \sqrt{1 -\beta_2^t} \frac{g_{t, j}^{(i)}}{\sqrt{v_{t, j}^{(i)}}}  \right)\mathds{1}(sign([\nabla_i f(x_t)]_j) \neq sign(g_{t,j}^{(i)})) \nonumber
\end{align}

Relying on the following inequalities: $\sqrt{v_t} \leq G$ and $1-\beta_2^t > 1-\beta_2$. 
\\
Taking expectation, we have the following:
\begin{align}
\mathbb{E}[T_1] &\leq - \eta_t \sum_{i=1}^h \sum_{j=1}^{d_i} \frac{\sqrt{1 - \beta_2}}{G}  \mathbb{E}\left[[\nabla_i f(x_t)]_j g_{t, j}^{(i)}\right] \nonumber \\
&\qquad \qquad - \eta_t \sum_{i=1}^h \sum_{j=1}^{d_i} \frac{\sqrt{1 - \beta_2}}{G} \mathbb{E}\left[ \left([\nabla_i f(x_t)]_j  g_{t, j}^{(i)}  \right)\mathds{1}(sign([\nabla_i f(x_t)]_j) \neq sign(g_{t,j}^{(i)}))\right] \nonumber \\
\mathbb{E}[T_1] &\leq - \eta_t \sum_{i=1}^h \sum_{j=1}^{d_i} \frac{\sqrt{1 - \beta_2}}{G}  \mathbb{E}\left[[\nabla_i f(x_t)]_j g_{t, j}^{(i)}\right] \nonumber \\
&\qquad \qquad + \eta_t \sum_{i=1}^h \sum_{j=1}^{d_i} \sqrt{1 - \beta_2} \mathbb{E}\left[ \left([\nabla_i f(x_t)]_j   \right)|\mathbb{P}(sign([\nabla_i f(x_t)]_j) \neq sign(g_{t,j}^{(i)}))\right]  
\label{eq:adam-eq-4}
\end{align}

similarly what is shown in signsgd, we bound the probability by first relaxing the condition, then applying Markov's and then Jensen's inequality:
\begin{align}
\mathbb{P}(sign([\nabla_i f(x_t)]_j) \neq sign(g_{t,j}^{(i)})) &\leq \mathbb{P}\left(|[\nabla_i f(x_t)]_j - g_{t,j}^{(i)}| \geq |g_{t,j}^{(i)}|\right) \nonumber \\
&\leq \frac{\mathbb{E}\left[|[\nabla_i f(x_t)]_j - g_{t,j}^{(i)}|\right]}{|[\nabla_i f(x_t)]_j|} \nonumber \\
&\leq \frac{\sqrt{\mathbb{E}\left[([\nabla_i f(x_t)]_j - g_{t,j}^{(i)})^2\right]}}{|[\nabla_i f(x_t)]_j|} \nonumber \\
&= \frac{\tilde{\sigma}_{t,i}}{|[\nabla_i f(x_t)]_j|} \nonumber \\
&\leq \frac{\tilde{\sigma}_{i}}{\sqrt{n}|[\nabla_i f(x_t)]_j|} \nonumber
\end{align}
where the last inequality is from the fact that $\tilde{\sigma}_{t,j}$ is the minibatch variance at time $t$ with batch size $n$. Substituting this into our derivation of $T_1$
\begin{align}
\mathbb{E}[T_1] &\leq - \eta_t \frac{\sqrt{1 - \beta_2}}{G}  ||\nabla f(x_t)||^2 + \eta_t \sqrt{1 - \beta_2} \sum_{i=1}^h \sum_{j=1}^{d_i} \frac{\tilde{\sigma}_{i}}{\sqrt{n}} \nonumber
\end{align}
and replacing this with our definition of $T_1$ in Eq. (\ref{eq:adam-eq-1}) we get
\begin{align}
\mathbb{E}[f(x_{t+1})] &\leq f(x_t) - \eta_t \frac{\sqrt{1 - \beta_2}}{G} ||\nabla f(x_t)||^2 + \eta_t \sqrt{1 - \beta_2} \frac{||\tilde{\sigma}||_1}{\sqrt{n}} + \frac{||L||_1\eta_t^2d}{2(1-\beta_2)}. \label{eq:adam-eq-3}
\end{align}
We then arrive at the final bound by summing Eq. (\ref{eq:adam-eq-3}) to step $T$ and cancelling consecutive terms via the telescoping sum, followed by rearranging and then multiplying through by $\frac{G}{T \eta_t \sqrt{1 - \beta_2}}$
\begin{align}
\mathbb{E}[f(x_{T+1})] &\leq f(x_1) - \frac{\eta_t \sqrt{1 - \beta_2}}{G}\sum_{t=1}^T ||\nabla f(x_t)||^2 + T \eta_t \sqrt{1 - \beta_2} \frac{||\tilde{\sigma}||_1}{\sqrt{n}} + \frac{T||L||_1\eta_t^2d}{2(1-\beta_2)}. \nonumber \\
\frac{\eta_t \sqrt{1 - \beta_2}}{G}\sum_{t=1}^T ||\nabla f(x_t)||^2 &\leq f(x_1) - f(x^*) + T \eta_t \sqrt{1 - \beta_2} \frac{||\tilde{\sigma}||_1}{\sqrt{n}} + \frac{T||L||_1\eta_t^2d}{2(1-\beta_2)} \nonumber \\
\frac{1}{T}\sum_{t=1}^T ||\nabla f(x_t)||^2 &\leq G\left(\frac{f(x_1) - f(x^*)}{T \eta_t \sqrt{1 - \beta_2}} + \frac{||\tilde{\sigma}||_1}{\sqrt{n}} + \frac{||L||_1 \eta_t d}{2(1 - \beta_2)^{\frac{3}{2}}}\right) \nonumber
\end{align}

Taking $\eta_t = \eta = \sqrt{\frac{2(f(x_1) - f(x^*))}{T||L||_1(1 - \beta_2)d}}$ and letting $n = T$ as is similarly done in \citep{you2019lamb}, we can recover a bound that, up to some constants, is similar to the bound for LAMB:
\begin{align}
\mathbb{E}[||\nabla f(x_t)||^2] &\leq \mathcal{O}\left(G\left(\frac{f(x_1) - f(x^*)}{T \sqrt{\frac{2(f(x_1) - f(x^*))}{T||L||_1(1 - \beta_2)d}} \sqrt{1 - \beta_2}} + \frac{||\tilde{\sigma}||_1}{\sqrt{n}} + \frac{||L||_1 \sqrt{\frac{2(f(x_1) - f(x^*))}{T||L||_1(1 - \beta_2)d}} d}{2(1 - \beta_2)^{\frac{3}{2}}}\right)\right) \nonumber \\
&= \mathcal{O}\left(G\left(\frac{1}{2}\sqrt{\frac{2(f(x_1) - f(x^*)) ||L||_1 d}{T}} + \frac{||\tilde{\sigma}||_1}{\sqrt{n}} + \frac{1}{2(1 - \beta_2)^2}\sqrt{\frac{2(f(x_1) - f(x^*)) ||L||_1 d}{T}}\right)\right) \nonumber \\
&= \mathcal{O}\left(G \left(1 + \frac{1}{(1 - \beta_2)^2}\right)\sqrt{\frac{2(f(x_1) - f(x^*))||L||_1 d}{T}} + \frac{G||\tilde{\sigma}||_1}{\sqrt{T}}\right) \nonumber
\end{align}

\end{proof}

\section{Additional experiment details}\label{appendix:experiment-details}

\subsection{ResNet-50 training benchmark}

All experiments were run on Google TPUs \citep{jouppi2017datacenter}. We typically trained on TPUv2-256 or TPUv3-128 in order to accommodate the 32,768 batch size. 
The ResNet-50 experiments used Jax \citep{jax2018github} using the Flax library, with code released \href{https://github.com/anonymized}{here}. The BERT experiments were run using TensorFlow \citep{abadi2016tensorflow}  version 1.15.
We used the standard train/validation split from the previous literature and MLPerf competition.

For ImageNet, we used the following sequence of TensorFlow functions for pre-processing:\footnote{Full code available at \url{https://git.io/JtgtE}}
\begin{lstlisting}
tf.image.sample_distorted_bounding_box
tf.image.decode_and_crop_jpeg
tf.image.resize
tf.image.random_flip_left_right
tf.image.convert_image_dtype
\end{lstlisting}

\subsection{BERT pre-training}\label{appendix:bert_experiment_details}

We used the same experimental setup as the official BERT codebase\footnote{\url{https://github.com/google-research/bert}} and the standard train/test split from the previous literature. This matches the experimental setup of \citet{you2019lamb}. We trained on Google TPUs, using TPUv3-256 or TPUv3-512 for the 32,768 batch size experiments, and TPUv3-1024 for the 65,536 batch size experiments.

We trained the two pretraining objectives on the combined Wikipedia and Books corpus \citep{ZhuEtAl2015bookcorpus} datasets (2.5B and 800M words, respectively). We used sequence lengths of 128 and 512, respectively, for the pretraining tasks. We ran the fine-tuning phase on the SQuaD v1.1 question answering task. In order to match \citet{you2019lamb}, we report the F1-score on the dev set as the target metric. We followed the fine-tuning protocol described in the LAMB optimizer setup and did not perform any additional tuning for fine-tuning.

We tuned Adam hyperparameters using quasi-random search \citep{bousquet2017critical} in a simple search space. Hyperparameters included learning rate $\eta$, $\beta_1$, $\beta_2$, the polynomial power for the learning rate warmup $p_{warmup}$, and weight decay $\lambda$. We fixed the $\epsilon$ in Adam to $\num{e-11}$ for all BERT experiments. See Appendix~\ref{appendix:adam_bert_hparams} for the search spaces. We selected the best trial using the masked language model accuracy over 10k examples from the training set. The number of training steps for each of the phases, as well as the warmup steps are identical to \citet{you2019lamb} and are listed in Appendix~\ref{appendix:adam_bert_hparams}.
Each phase of pretraining used completely independent Adam hyperparameters. We found the final hyperparameters within 30 trials of random search for each of the phases, except for the second phase of 65,536 batch size which used 130 trials. 

\begin{figure}[t]
    \centering
    \includegraphics[width=\linewidth]{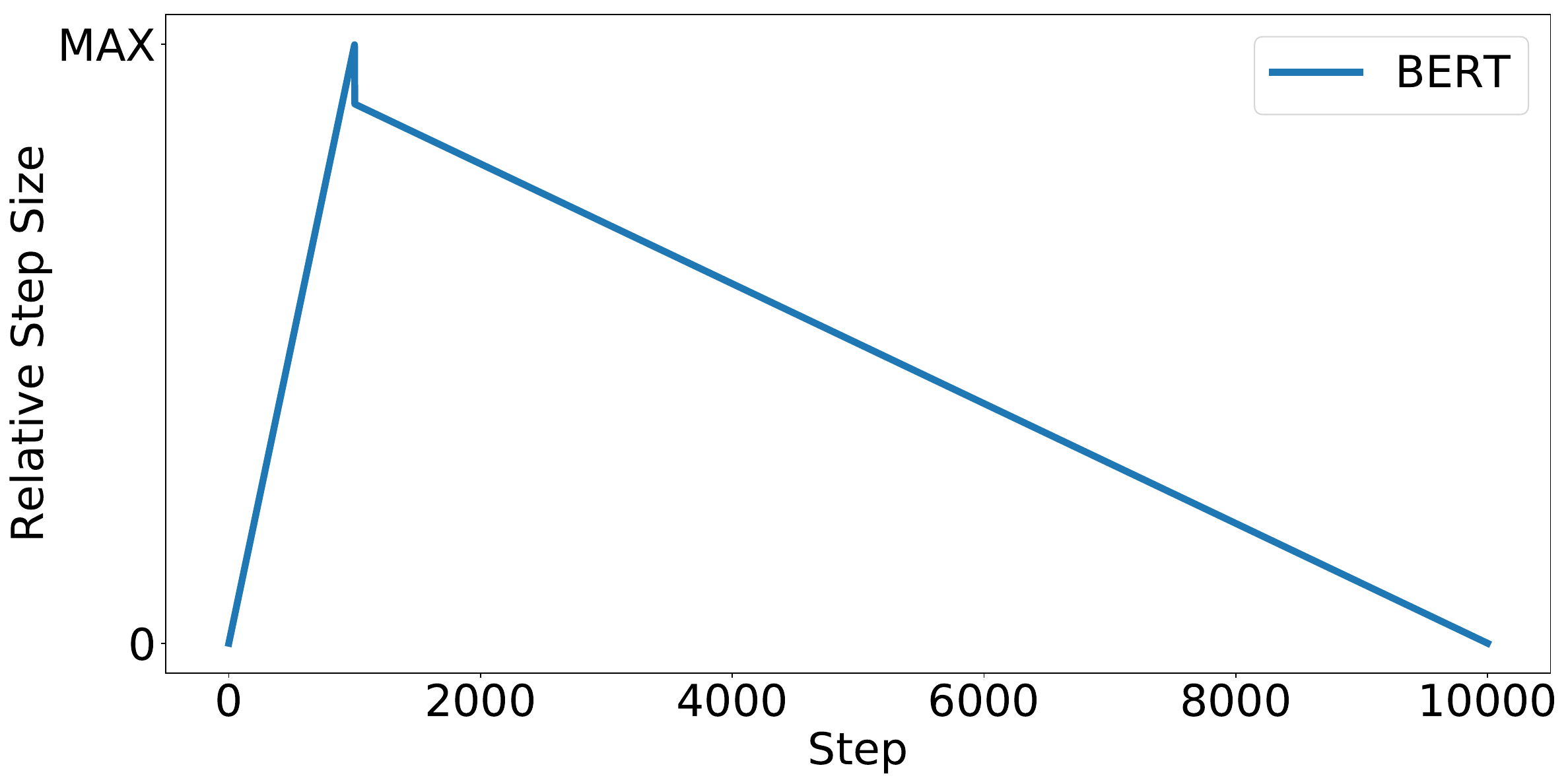}
    \caption{An illustration of the sudden drop in the BERT learning rate schedule in the official codebase.}
    \label{fig:mlperf_bert_schedules}
\end{figure}

\begin{figure}[t]
    \centering
    \includegraphics[width=\linewidth]{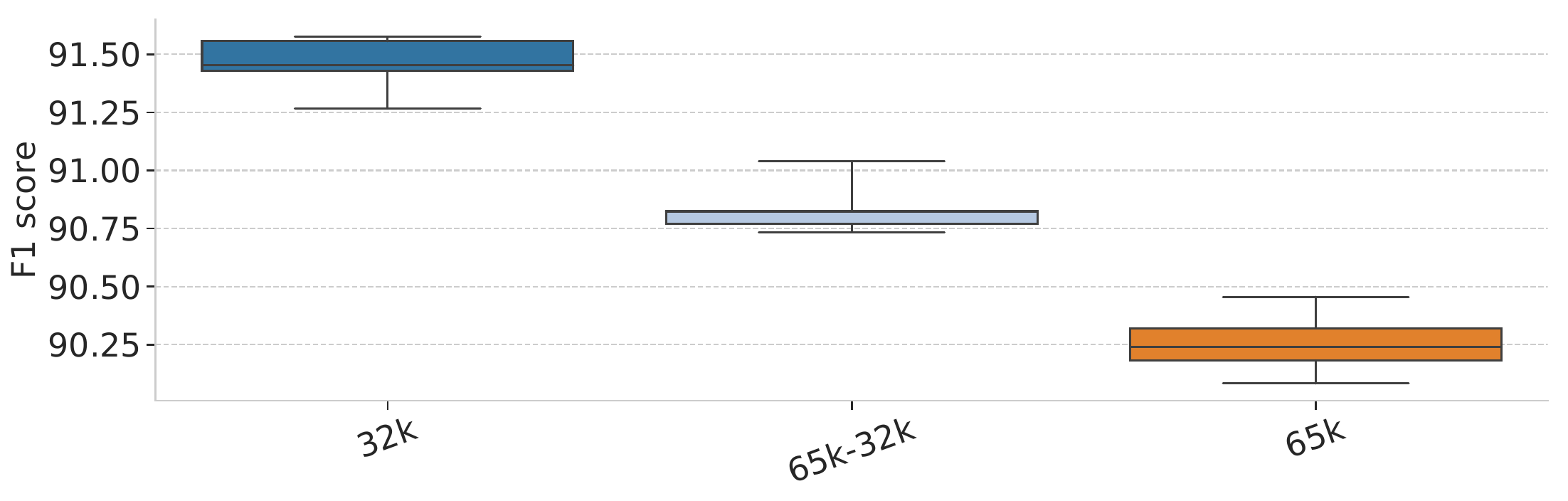}
    \caption{6 finetuning runs starting from the same pretraining checkpoint to show the stability of our results, at each of the 32,768, mixed 65,536-32,768, and 65,536 batch size settings.}
\end{figure}

\section{Nesterov ablations}\label{appendix:nesterov-ablations}
\begin{figure}[t]
    \centering
    \includegraphics[width=\linewidth]{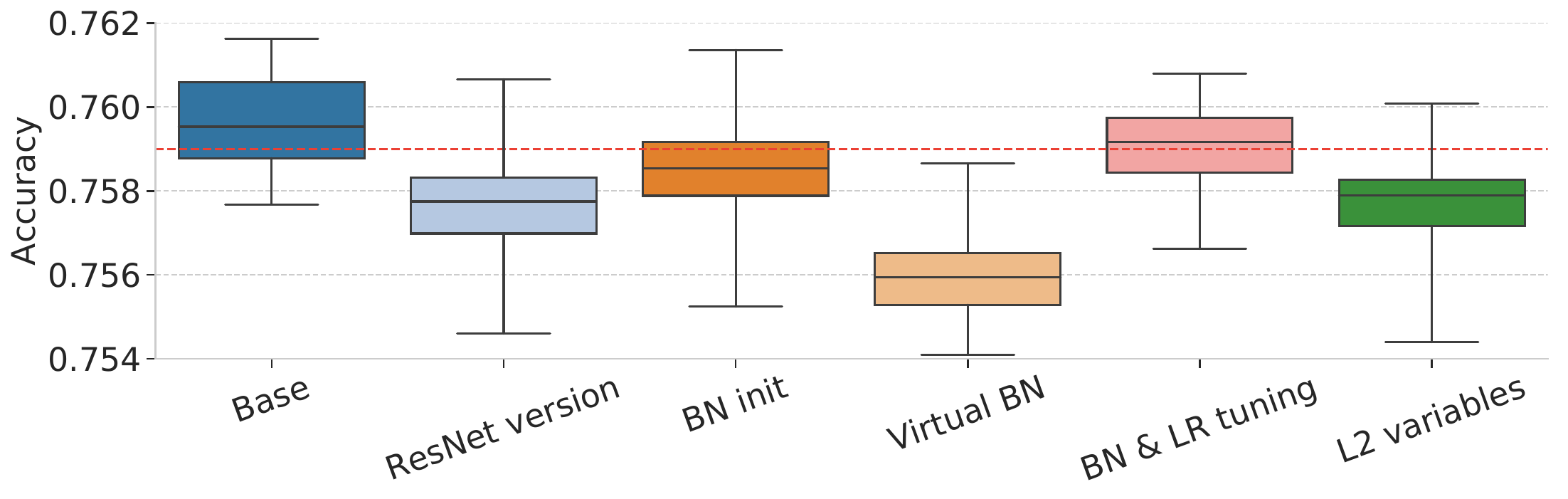}
    \caption{Distributions over 50 training runs for each ablation study around our best Nesterov momentum configuration (Configuration A). The dotted red line is at the target accuracy of 75.9\%, and the boxes show the min, max, and quartiles of the distribution of accuracies over the 50 training runs.}
    \label{fig:one_off_ablations_nesterov}
\end{figure}

To explore the sensitivity of our best Nesterov momentum configuration (Configuration A), we ablated several elements of the experiment pipeline, one at a time, and tested their impact on performance. Figure~\ref{fig:one_off_ablations_nesterov} shows the results of these experiments. ``Base'' refers to Nesterov momentum Configuration A (Table~\ref{table:nesterov_config}). ``ResNet version'' is the same point as ``Base'' but with ResNet version 1.0 instead of version 1.5. ``BN init'' is the same point as ``Base'' but with $\gamma_0 = 1.0$ instead of 0.4138. ``Virtual BN'' is the same point as ``Base'' but with a virtual batch size of 256 instead of 64, which is the largest that fits in a single TPUv3 core. ``BN \& LR tuning'' is Configuration B (Table~\ref{table:nesterov_config}), the same point as ``Base'' but with $p_{decay}, t_{warmup}, \eta_0, \rho, \epsilon$ set to their values in the LARS pipeline. Finally, ``L2 variables'' is the same point as ``Base'' but where the L2 regularization is applied to all variables.
The only ablation whose median over 50 seeds continues to beat the target 75.9\% accuracy (noted by the dotted red line) is ``BN \& LR tuning'', with the rest having between 0.1\%-0.3\% drops in median accuracy.

\section{Hyperparameter tuning}\label{appendix:hparam_tuning}

\subsection{Nesterov momentum training speed on ResNet-50}
\label{appendix:nesterov_config}

We considered two configurations of Nesterov hyperparameters: Configuration A, where we tuned a wide set of hyperparameters in the experiment pipeline, and Configuration B, where we reverted the less impactful hyperparameters to the same values as the LARS baseline (or in the case of $p_\text{warmup}$, a simpler value). We included Configuration B in order to demonstrate the minimal set of changes to the baseline necessary to still reach the target accuracy. The hyperparameter values for these configurations can be found in Table~\ref{table:nesterov_config}. %

\begin{table}[t]
\centering
\setlength{\extrarowheight}{3.5pt}
\begin{tabular}{|c|c|c|c|}
\hline
 & Configuration A & Configuration B & LARS \\ \hline
$t_\text{warmup}$ & 638 & 706 & 706 \\ \hline
$p_\text{warmup}$ & 2.497 & 2.0 & 1.0 \\ \hline
$p_\text{decay}$ & 1.955 & 2.0 & 2.0 \\ \hline
$\rho$ & 0.94 & 0.9 & 0.9\\ \hline
$\epsilon$ & $\num{4e-6}$ & $\num{e-5}$ & $\num{e-5}$  \\ \hline
$\eta_\text{peak}$ & 7.05 & 7.05 & 29.0 \\ \hline
$\eta_\text{final}$ & $\num{6e-6}$ & $\num{6e-6}$ & $\num{e-4}$ \\ \hline
$1 - \mu$ & 0.02397 & 0.02397 & 0.071 \\ \hline
$\lambda$ & $\num{5.8e-5}$ & $\num{5.8e-5}$ & \num{e-4} \\ \hline
$\tau$ & 0.15 & 0.15 & 0.10 \\ \hline
$\gamma_0$ & 0.4138 & 0.4138 & 0.0 \\ \hline
\end{tabular}
\caption{Nesterov momentum Configurations A and B.}\label{table:nesterov_config}
\end{table}

\subsection{Adam on BERT}
\label{appendix:adam_bert_hparams}
The search space used to tune Adam on BERT for all phases of the pipeline can be found in Table~\ref{table:bert_adam_tuning_space}, which yielded our best Adam results on BERT in Table~\ref{table:bert_results_params}.

\begin{table}[t]
\label{table:search_space_bert}
\centering
\setlength{\extrarowheight}{3.5pt}
\begin{tabular}{|c|c|c|}
\hline
Hyperparameter & Range & Scaling \\ \hline
$p$ & $\{1, 2\}$ & Discrete \\ \hline
$\eta$ & $\srange{e-5}{1.0}$ & Log \\ \hline
$1 - \beta_1$ & $\srange{e-2}{0.5}$ & Log \\ \hline
$1 - \beta_2$ & $\srange{e-2}{0.5}$ & Log \\ \hline
$\lambda$ & $\srange{e-3}{10}$ & Log \\ \hline
\end{tabular}
\caption{The search space used to tune Adam on BERT for all phases of the pipeline. $\lambda$ refers to weight decay and $p$ refers to the polynomial power in the learning rate schedule for both the warmup and decay phases.}
\label{table:bert_adam_tuning_space}
\end{table}

\begin{table*}[t!]
\centering
 \begin{tabular}{|c|c|c|c|c|c|c|c|c|c|}
\hline
Batch size & Phase & Seq len & Warmup & Train & Learning & $\beta_1$ & $\beta_2$ & $\lambda$ & $p$ \\
& & & steps & steps & rate & & & & \\
\hline
32,768 &  1 & 128 & 3,125 & 14,063 & \num{5.9415e-4} & 0.934271 & 0.989295  & 0.31466 & 1 \\
\hline
32,768 &  2 & 512 & 781 & 1,562 & \num{2.8464e-4} & 0.963567 & 0.952647  & 0.31466 & 1 \\
\hhline{|=|=|=|=|=|=|=|=|=|=|}
65,536 &  1 & 128 & 2,000 & 7,037 & \num{1.3653e-3} & 0.952378 & 0.86471  & 0.19891 & 2 \\
\hline
32,768 &  2 & 512 & 781 & 1,562 & \num{2.8464e-4} & 0.952647 & 0.963567  & 0.19891 & 2\\
\hline
65,536 &  2 & 512 & 390 & 781 & \num{6.1951e-5} & 0.65322 & 0.82451  & 0.19891 & 2 \\
\hline
\end{tabular}
\caption{Best hyperparameters from tuning Adam on BERT-Large pretraining. $\lambda$ refers to weight decay and $p$ refers to the polynomial power in the learning rate schedule for both the warmup and decay phases. All trials used $\epsilon=\num{e-11}$.}
\label{table:bert_results_params}
\end{table*}

\begin{table*}[t]
\centering
\setlength{\extrarowheight}{3.5pt}
\begin{tabular}{|c|c|c|c|c|c|}
\hline
Weights Optimizer & Bias/BN Optimizer & Name & Initial Range & Final Range & Best \\ \hline
Nesterov & Nesterov & $\eta$ & np.logspace(-.5, .5, 10) & $\srange{0.8}{3}$ & 1.173 \\ \hline
Nesterov & Nesterov & $\lambda$ & np.logspace(-4, -3, 10) & $\srange{3e-4}{e-3}$ & $\num{3.026e-4}$ \\ \hhline{|=|=|=|=|=|=|}
LARS & \specialcell{Heavy-ball\\momentum} & $\eta$ & np.logspace(0, 2, 10) & $\srange{10}{40}$ & 14.49 \\ \hline
LARS & \specialcell{Heavy-ball\\momentum} & $\lambda$ & np.logspace(-5, -2, 10) & $\srange{5e-5}{2e-4}$ & $\num{1.708e-4}$ \\ \hhline{|=|=|=|=|=|=|}
LARS & LARS & $\eta$ & $\srange{e0}{30}$ & $\srange{10}{30}$ & 14.18 \\ \hline
LARS & LARS & $\lambda$ & $\srange{e-4}{e-1}$ & $\srange{5e-5}{5e-4}$ & $\num{5.278e-5}$ \\ \hhline{|=|=|=|=|=|=|}
Adam ($\epsilon=10^{-8}$) & Adam ($\epsilon=10^{-8}$) & $\eta$ & $\srange{e-3}{e0}$ & $\srange{4e-3}{2e-2}$ & 0.004596 \\ \hline
Adam ($\epsilon=10^{-8}$) & Adam ($\epsilon=10^{-8}$) & $\lambda$ & $\srange{e-2}{4e0}$ & $\srange{2e-1}{1}$ & 0.6182 \\ \hhline{|=|=|=|=|=|=|}
Adam ($\epsilon=10^{-6}$) & Adam ($\epsilon=10^{-6}$) & $\eta$ & np.logspace(-3, 0, 10) & $\srange{3e-3}{e-2}$ & $\num{3.332e-3}$ \\ \hline
Adam ($\epsilon=10^{-6}$) & Adam ($\epsilon=10^{-6}$) & $\lambda$ & np.logspace(-2, 0.5, 6) & $\srange{0.5}{2}$ & 1.055 \\ \hhline{|=|=|=|=|=|=|}
LAMB & LAMB & $\eta$ & np.logspace(-4, 0, 30) & $\srange{4e-3}{5e-2}$ & 0.01134 \\ \hline
LAMB & LAMB & $\lambda$ & np.logspace(-5, -2, 4) & $\srange{1e-2}{0.1}$ & 0.02657 \\ \hhline{|=|=|=|=|=|=|}
LAMB & Adam ($\epsilon=10^{-8}$) & $\eta$ & $\srange{e-3}{1}$ & $\srange{e-2}{8e-2}$ & 0.02569 \\ \hline
LAMB & Adam ($\epsilon=10^{-8}$) & $\lambda$ & $\srange{e-2}{4}$ & $\srange{1}{8}$ & 2.500 \\ \hhline{|=|=|=|=|=|=|}
LAMB & Adam ($\epsilon=10^{-6}$) & $\eta$ & np.logspace(-3, 0, 10) & $\srange{e-2}{8e-2}$ & 0.03378 \\ \hline
LAMB & Adam ($\epsilon=10^{-6}$) & $\lambda$ & np.logspace(-2, 0.5, 6) & $\srange{1}{8}$ & 4.197 \\ \hline
\end{tabular}
\caption{Search spaces used for the 6,000 step, cosine learning rate schedule experiments. All hyperparameters were tuned on a logarithmic scale, except for those which define a discrete sequence of points to evaluate such as ``np.logspace''.}
\label{table:cosine_hparam_tunings}
\end{table*}

\subsection{Less stringent step budget on ResNet-50}
\label{appendix:cosine_details}
All trials used a cosine decay learning rate schedule and tuned the initial learning rate $\eta$ and L2 regularization or weight decay parameter\footnote{As suggested in \citet{you2019lamb}, we used L2 regularization for LARS and weight decay for LAMB. For consistency, we used L2 regularization for Nesterov momentum (which is more analogous to LARS) and weight decay for Adam (which is more analogous to LAMB).} $\lambda$ according to Table~\ref{table:cosine_hparam_tunings}. We used 50 or more trials to search in the ``Initial Range'' and then 25 trials to search in the refined ``Final Range.'' Finally, we ran the best point from the latter for 5 random seeds. When LARS or LAMB were used alongside a different optimizer for the batch normalization and ResNet-50 bias parameters, we set $\lambda = 0$ on the batch normalization and ResNet-50 bias parameters. When LAMB was used all parameters, the majority of trials diverged during training -- it took \textbf{67 trials} to get 25 trials that did not NaN during training. Our trial budgets refer to the number of feasible trials, i.e. trials that do not diverge during training.

\subsection{Nesterov ResNet50 search space chronology}
\label{appendix:historical_search_spaces_nesterov_resnet50}

Below we list the sequence of search spaces we used to arrive at our final values in Table~\ref{table:nesterov_config}. Given that the final results reported in papers are rarely found in a single iteration of experiments, we believe that it is important to document the full journey to arriving at our results. 

Note that although we tuned a wide range of hyperparameters to match the LARS result with Nesterov momentum, we later realized that many of these hyperparameters could be reverted to the values from the LARS pipeline (see Table~\ref{table:nesterov_config}).
We started tuning with a training budget of 2,815 steps, which is the number of steps in the MLPerf 0.6 submission. We sometimes would decrease this to 2,658 steps to test how decreasing the training budget would affect tuning performance, before eventually moving to the 2,512 steps used to generate the results in the main text.

\begin{table}[t]
\centering
\setlength{\extrarowheight}{3.5pt}
\begin{tabular}{|c|c|c|}
\hline
 & Range & Scaling \\ \hline
$\eta_0$ & $\srange{e-3}{50.0}$ & Log \\ \hline
$\eta_\text{decay\_factor}$ & $\{\num{e-4}, \num{e-3}, \num{e-2}, \num{e-1}\}$ & Discrete \\ \hline
$1 - \mu$ & $\srange{e-3}{1.0}$ & Log \\ \hline
$\lambda$ & $\srange{e-5}{e-1}$ & Log \\ \hline
$\tau$ & $\srange{e-2}{2e-1}$ & Linear \\ \hline
\end{tabular}
\caption{First search space of the Nesterov tuning journey. The search spaces were mostly by informed guesses by the authors. $\lambda$ refers to weight decay, which is applied to all variables. Tuned for 251 trials. Trained for 2,815 steps (``72 epochs'' as defined by MLPerf epoch calculations). We used a linear learning rate decay schedule that decays for all training steps, starting from $\eta_0$ and ending at $\eta_0 \times \eta_{decay\_factor}$. Virtual batch size 128.}
\label{table:imagenet_mlperf_nesterov_vizier_2815}
\end{table}

\begin{table}[t]
\centering
\setlength{\extrarowheight}{3.5pt}
\begin{tabular}{|c|c|c|}
\hline
 & Range & Scaling \\ \hline
$\eta_0$ & $\srange{e-3}{50.0}$ & Log \\ \hline
$\eta_\text{decay\_factor}$ & $\{\num{e-4}, \num{e-3}, \num{e-2}, \num{e-1}\}$ & Discrete \\ \hline
$1 - \mu$ & $\srange{e-3}{1.0}$ & Log \\ \hline
$\lambda$ & $\srange{e-5}{e-1}$ & Log \\ \hline
$\tau$ & $\srange{e-2}{2e-1}$ & Linear \\ \hline
\end{tabular}
\caption{Same as Table~\ref{table:imagenet_mlperf_nesterov_vizier_2815} but trained for 2,658 steps (``68 epochs'' as defined by MLPerf epoch calculations) for 50 trials.}
\end{table}

\begin{table}[t]
\centering
\setlength{\extrarowheight}{3.5pt}
\begin{tabular}{|c|c|c|}
\hline
 & Range & Scaling \\ \hline
$\eta_0$ & $\srange{e-1}{20.0}$ & Log \\ \hline
$\eta_\text{decay\_factor}$ & $\{\num{e-5}, \num{e-4}, \num{e-3}\}$ & Discrete \\ \hline
$t_\text{decay}$ & $\srange{2392}{2,658}$ & Linear \\ \hline
$1 - \mu$ & $\srange{e-3}{1.0}$ & Log \\ \hline
$\lambda$ & $\srange{e-5}{2e-1}$ & Log \\ \hline
$\tau$ & $\srange{e-2}{2e-1}$ & Linear \\ \hline
\end{tabular}
\caption{$\lambda$ refers to weight decay, which is now not applied to the bias and batch normalization variables. 50 trials. Trained for 2,658 steps. Linear learning rate decay schedule that decays for $t_\text{decay}$ steps, starting from $\eta_0$ and ending at $\eta_0 \times \eta_{decay\_factor}$. Virtual batch size 128.}
\end{table}

\begin{table}[t]
\centering
\setlength{\extrarowheight}{3.5pt}
\begin{tabular}{|c|c|c|}
\hline
 & Range & Scaling \\ \hline
$\eta_\text{peak}$ & $\srange{e-1}{32.0}$ & Log \\ \hline
$\eta_\text{decay\_factor}$ & $\{\num{e-5}, \num{e-4}, \num{e-3}\}$ & Discrete \\ \hline
$t_\text{decay}$ & $\srange{2392}{2,658}$ & Linear \\ \hline
$1 - \mu$ & $\srange{e-4}{e-1}$ & Log \\ \hline
$\lambda$ & $\srange{e-4}{e-1}$ & Log \\ \hline
$\tau$ & $\srange{5e-2}{0.15}$ & Linear \\ \hline
\end{tabular}
\caption{$\lambda$ refers to weight decay, which is not applied to the bias and batch normalization variables. 50 trials. Trained for 2,658 steps. Linear warmup for 500 steps followed by a quadratic decay, which decays until step $t_\text{decay}$, and then is constant at the final learning rate $\eta_0 \times \eta_{decay\_factor}$. Virtual batch size 128. We increased the max learning rate based off the larger learning rates used by LARS. \textbf{We also ran two additional studies which were the same except with 250 and 977 warmup steps.}}
\end{table}

\begin{table}[t]
\centering
\setlength{\extrarowheight}{3.5pt}
\begin{tabular}{|c|c|c|}
\hline
 & Range & Scaling \\ \hline
$\eta_\text{peak}$ & $\srange{e-1}{32.0}$ & Log \\ \hline
$\eta_\text{decay\_factor}$ & $\srange{3e-5}{3e-4}$ & Log \\ \hline
$t_\text{decay}$ & $\srange{2533}{2,815}$ & Linear \\ \hline
$1 - \mu$ & $\srange{e-4}{e-1}$ & Log \\ \hline
$\lambda$ & $\srange{e-4}{e-1}$ & Log \\ \hline
$\tau$ & $\srange{5e-2}{0.15}$ & Linear \\ \hline
\end{tabular}
\caption{$\lambda$ refers to weight decay, which is not applied to the bias and batch normalization variables. 50 trials. Trained for 2,815 steps. Linear warmup for 500 steps followed by a quadratic decay, which decays until step $t_\text{decay}$, and then is constant at the final learning rate $\eta_0 \times \eta_{decay\_factor}$. Virtual batch size 128.}
\end{table}

\begin{table}[t]
\centering
\setlength{\extrarowheight}{3.5pt}
\begin{tabular}{|c|c|c|}
\hline
 & Range & Scaling \\ \hline
$\eta_\text{peak}$ & $\srange{e-1}{32.0}$ & Log \\ \hline
$\eta_\text{decay\_factor}$ & $\srange{3e-5}{3e-4}$ & Log \\ \hline
$t_\text{decay}$ & $\srange{2533}{2,815}$ & Linear \\ \hline
$1 - \mu$ & $\srange{5e-3}{e-1}$ & Log \\ \hline
$\lambda$ & $\srange{e-2}{e-1}$ & Log \\ \hline
$\tau$ & $\srange{5e-2}{0.15}$ & Linear \\ \hline
\end{tabular}
\caption{$\lambda$ refers to weight decay, which is not applied to the bias and batch normalization variables. 50 trials. Trained for 2,815 steps. Linear warmup for 500 steps followed by a quadratic decay, which decays until step $t_\text{decay}$, and then is constant at the final learning rate $\eta_0 \times \eta_{decay\_factor}$. Virtual batch size 128.}
\label{table:imagenet_mlperf_nesterov_viz2_500w_2815}
\end{table}

\begin{table}[t]
\centering
\setlength{\extrarowheight}{3.5pt}
\begin{tabular}{|c|c|c|}
\hline
 & Range & Scaling \\ \hline
$\eta_\text{peak}$ & $\srange{e-1}{32.0}$ & Log \\ \hline
$\eta_\text{decay\_factor}$ & $\srange{3e-5}{3e-4}$ & Log \\ \hline
$t_\text{decay}$ & $\srange{2533}{2,815}$ & Linear \\ \hline
$1 - \mu$ & $\srange{5e-3}{e-1}$ & Log \\ \hline
$\lambda$ & $\srange{e-2}{e-1}$ & Log \\ \hline
$\tau$ & $\srange{5e-2}{0.15}$ & Linear \\ \hline
\end{tabular}
\caption{The same as Table~\ref{table:imagenet_mlperf_nesterov_viz2_500w_2815} except with virtual batch size 64.}
\label{table:imagenet_mlperf_nesterov_viz_vbn64_2815}
\end{table}

\begin{table}[t]
\centering
\setlength{\extrarowheight}{3.5pt}
\begin{tabular}{|c|c|c|}
\hline
 & Range & Scaling \\ \hline
$\eta_\text{peak}$ & \specialcell{$\{\{10^\alpha, 2\times 10^\alpha, ..., 9\times 10^\alpha\}$ \\ $\forall \alpha \in \{-3, ... 2\}\} + \{100,\}$} & Discrete \\ \hline
$\eta_\text{decay\_factor}$ & $\num{8.144e-5}$ & -- \\ \hline
$t_\text{decay}$ & 2250 & -- \\ \hline
$1 - \mu$ & 0.02397 & -- \\ \hline
$\lambda$ & 0.009992 & -- \\ \hline
$\tau$ & 0.07786 & -- \\ \hline
\end{tabular}
\caption{$\lambda$ refers to weight decay, which is not applied to the bias and batch normalization variables. Trained for 2,815 steps. Virtual batch size 64. Using the best hyperparameters from Table~\ref{table:imagenet_mlperf_nesterov_viz_vbn64_2815}, we swept over the peak learning rate in a discrete set of ten values per order of magnitude, \textbf{each for three random seeds}, to find the max stable learning rate.}
\label{table:imagenet_mlperf_nesterov_lrtune_2815}
\end{table}

\begin{table}[t]
\centering
\setlength{\extrarowheight}{3.5pt}
\begin{tabular}{|c|c|c|}
\hline
 & Range & Scaling \\ \hline
$\eta_\text{peak}$ &4.118 & -- \\ \hline
$\eta_\text{decay\_factor}$ & $\num{8.144e-5}$ & -- \\ \hline
$t_\text{decay}$ & 2250 & -- \\ \hline
$1 - \mu$ & 0.02397 & -- \\ \hline
$\lambda$ & \specialcell{$\{\{0.5\times 10^\alpha, 10^\alpha, ...\}$ \\ $\forall \alpha \in \{-3, ... 0\}\} + \{1.0,\}$} & Discrete \\ \hline
$\tau$ & 0.07786 & -- \\ \hline
\end{tabular}
\caption{$\lambda$ refers to weight decay, which is not applied to the bias and batch normalization variables. Trained for 2,815 steps. Virtual batch size 64. Using the best hyperparameters from Table~\ref{table:imagenet_mlperf_nesterov_viz_vbn64_2815}, we swept over the weight decay in a discrete set of twenty values per order of magnitude, to test how high the regularization has to be in this region of hyperparameter space.}
\label{table:imagenet_mlperf_nesterov_wdtune_2815}
\end{table}

\begin{table}[t]
\centering
\setlength{\extrarowheight}{3.5pt}
\begin{tabular}{|c|c|c|}
\hline
 & Range & Scaling \\ \hline
$\eta_\text{peak}$ & 4.118 & -- \\ \hline
$\eta_\text{decay\_factor}$ & $\num{8.144e-5}$ & -- \\ \hline
$t_\text{decay}$ & 2250 & -- \\ \hline
$1 - \mu$ & 0.02397 & -- \\ \hline
$\lambda$ & 0.009992 & -- \\ \hline
$\tau$ & 0.07786 & -- \\ \hline
$\rho$ & \specialcell{$\{0.0, 0.1, 0.3, 0.5, 0.6, 0.7,$ \\ $0.8, 0.9, 0.95, 0.995, 0.999\}$} & Discrete \\ \hline
$\epsilon$ & \specialcell{$\{\num{e-7}, \num{e-6}, \num{e-5}, \num{e-4},$ \\ $\num{e-3}, \num{e-2}, \num{e-1}\}$} & Discrete \\ \hline
\end{tabular}
\caption{$\lambda$ refers to weight decay, which is not applied to the bias and batch normalization variables. Trained for 2,815 steps. Virtual batch size 64. Using the best hyperparameters from Table~\ref{table:imagenet_mlperf_nesterov_viz_vbn64_2815}, we swept over batch normalization hyperparameters.}
\end{table}

\begin{table}[t]
\centering
\setlength{\extrarowheight}{3.5pt}
\begin{tabular}{|c|c|c|}
\hline
 & Range & Scaling \\ \hline
$\eta_\text{peak}$ & $\lbrack2.0, 8.0\rbrack$ & Log \\ \hline
$\eta_\text{decay\_factor}$ & $\srange{4e-5}{1.6e-4}$ & Linear \\ \hline
$t_\text{decay}$ & $\srange{2100}{2400}$ & Linear \\ \hline
$1 - \mu$ & $\srange{0.012}{0.04}$ & Log \\ \hline
$\lambda$ & $\srange{7e-3}{7e-2}$ & Log \\ \hline
$\tau$ & $\srange{0.04}{0.1}$ & Linear \\ \hline
$\rho$ & $\lbrack0.45, 0.55\rbrack$ & Linear \\ \hline
$\epsilon$ & $\srange{5e-6}{5e-5}$ & Linear \\ \hline
\end{tabular}
\caption{$\lambda$ refers to weight decay, which is not applied to the bias and batch normalization variables. 50 trials. Trained for 2,815 steps. Linear warmup for 500 steps followed by a quadratic decay, which decays until step $t_\text{decay}$, and then is constant at the final learning rate $\eta_0 \times \eta_{decay\_factor}$. Virtual batch size 64. Peak learning rate range was consolidated based off the results of Table~\ref{table:imagenet_mlperf_nesterov_lrtune_2815}. The weight decay range was consolidated based off the results of Table~\ref{table:imagenet_mlperf_nesterov_wdtune_2815}.}
\end{table}

\begin{table}[t]
\centering
\setlength{\extrarowheight}{3.5pt}
\begin{tabular}{|c|c|c|}
\hline
 & Range & Scaling \\ \hline
$t_\text{warmup}$ & $\srange{300}{800}$ & Linear \\ \hline
$p_\text{warmup}$ & $\lbrack0.7, 2.0\rbrack$ & Linear \\ \hline
$p_{decay}$ & 1.8 & -- \\ \hline
$\eta_0$ & $\srange{0.1}{1.0}$ & Log \\ \hline
$\eta_\text{peak}$ & $\srange{5.0}{9.0}$ & Log \\ \hline
$\eta_\text{final}$ & $\srange{e-5}{5e-5}$ & Log \\ \hline
$1 - \mu$ & 0.02397 & -- \\ \hline
$\lambda$ & $\num{5e-5}$ & -- \\ \hline
$\tau$ & 0.15 & -- \\ \hline
$\gamma_0$ & $\lbrack0.0, 0.6\rbrack$ & Linear \\ \hline
$\rho$ & 0.94 & -- \\ \hline
$\epsilon$ & $\num{4e-6}$ & -- \\ \hline
\end{tabular}
\caption{Here we switched $\lambda$ to refer to L2 regularization. We also began training for 2,512 steps, which is the final ``64 epochs'' used in the Nesterov results reported in the main text. Because of this more stringent step budget, we focused on the learning rate schedule. $t_\text{decay}$ was set to all remaining steps after the warmup was finished. Tuned for 229 trials. Virtual batch size 64.}
\label{table:momentum_vizier4_random}
\end{table}

\begin{table}[t]
\centering
\setlength{\extrarowheight}{3.5pt}
\begin{tabular}{|c|c|c|}
\hline
 & Range & Scaling \\ \hline
$t_{warmup}$ & 638 & -- \\ \hline
$p_\text{warmup}$ & $\lbrack1.5, 3.0\rbrack$ & Linear \\ \hline
$p_\text{decay}$ & $\lbrack1.5, 2.5\rbrack$ & Linear \\ \hline
$\eta_0$ & 0.12 & -- \\ \hline
$\eta_{peak}$ & 7.05 & -- \\ \hline
$\eta_\text{final}$ & $\srange{e-6}{5e-4}$ & Log \\ \hline
$1 - \mu$ &  0.02397 & -- \\ \hline
$\lambda$ & $\srange{5e-5}{1e-3}$ & Log \\ \hline
$\tau$ & 0.15 & -- \\ \hline
$\gamma_0$ & $\lbrack0.4, 1.0\rbrack$ & Linear \\ \hline
$\rho$ & 0.94 & -- \\ \hline
$\epsilon$ & $\num{4e-6}$ & -- \\ \hline
\end{tabular}
\caption{Here we began focusing more on the shape of the learning rate schedule, as well as retuning the L2 regularization. $\lambda$ refers to L2. Several values were picked from the best trial of Table~\ref{table:momentum_vizier4_random}. Trained for 2,512 steps steps. Tuned for 15 trials. Virtual batch size 64.}
\label{table:momentum_vizier5_random}
\end{table}

\begin{table}[t]
\centering
\setlength{\extrarowheight}{3.5pt}
\begin{tabular}{|c|c|c|}
\hline
 & Range & Scaling \\ \hline
$t_{warmup}$ & 638 & -- \\ \hline
$p_\text{warmup}$ & $\lbrack1.5, 3.0\rbrack$ & Linear \\ \hline
$p_\text{decay}$ & $\lbrack1.5, 2.5\rbrack$ & Linear \\ \hline
$\eta_0$ & 0.12 & -- \\ \hline
$\eta_{peak}$ & 7.05 & -- \\ \hline
$\eta_\text{final}$ & $\srange{e-6}{5e-4}$ & Log \\ \hline
$1 - \mu$ &  0.02397 & -- \\ \hline
$\lambda$ & $\srange{1e-5}{1e-4}$ & Log \\ \hline
$\tau$ & 0.15 & -- \\ \hline
$\gamma_0$ & $\lbrack0.4, 1.0\rbrack$ & Linear \\ \hline
$\rho$ & 0.94 & -- \\ \hline
$\epsilon$ & $\num{4e-6}$ & -- \\ \hline
\end{tabular}
\caption{Here we focus in more on tuning the L2 regularization. $\lambda$ refers to L2. Trained for 2,512 steps steps. Tuned for 37 trials. Virtual batch size 64.}
\end{table}

\begin{table}[t]
\centering
\setlength{\extrarowheight}{3.5pt}
\begin{tabular}{|c|c|c|}
\hline
 & Range & Scaling \\ \hline
$t_{warmup}$ & 638 & -- \\ \hline
$p_\text{warmup}$ & $\lbrack1.5, 3.0\rbrack$ & Linear \\ \hline
$p_\text{decay}$ & $\lbrack1.5, 2.5\rbrack$ & Linear \\ \hline
$\eta_0$ & 0.12 & -- \\ \hline
$\eta_{peak}$ & 7.05 & -- \\ \hline
$\eta_\text{final}$ & $\srange{e-6}{5e-4}$ & Log \\ \hline
$1 - \mu$ &  0.02397 & -- \\ \hline
$\lambda$ & $\srange{5e-5}{6e-5}$ & Linear \\ \hline
$\tau$ & 0.15 & -- \\ \hline
$\gamma_0$ & $\lbrack0.4, 1.0\rbrack$ & Linear \\ \hline
$\rho$ & 0.94 & -- \\ \hline
$\epsilon$ & $\num{4e-6}$ & -- \\ \hline
\end{tabular}
\caption{Again we dial in more on a tighter tuning range for the L2 regularization. $\lambda$ refers to L2. Trained for 2,512 steps steps. Tuned for 37 trials. Virtual batch size 64.}
\end{table}

\end{document}